\documentclass{llncs}

\usepackage[vlined,ruled]{algorithm2e_modif}
\usepackage[latin1]{inputenc}

\usepackage{amsfonts}
\usepackage{amssymb}
\usepackage{amsmath}
\usepackage{graphicx}

\newcommand{\sseq}{\subseteq}
\newcommand{\set}[1]{\{ #1 \}}
\newcommand{\size}[1]{ | #1 |}
\newcommand{\et}{\wedge}

\newcommand{\OO}{\ensuremath{\mathcal{O}}}
\renewcommand{\AA}{\ensuremath{\mathcal{A}}}
\newcommand{\CC}{\ensuremath{\mathcal{C}}} 
\newcommand{\db}{\ensuremath{Db}}
\newcommand{\txtrm}[1]{\textup{\sffamily #1}}
\newcommand{\concept}{\txtrm{Concepts}} 

\newcommand{\defnt}[1]{\textbf{#1}}

\begin{document}

\title{A Model for Managing  Collections of Patterns
}

\author{
Baptiste Jeudy \and
Christine Largeron \and
François Jacquenet
}

\institute{Laboratoire Hubert Curien, UMR CNRS 5516\\
       Univ. of St-Etienne, France}

\maketitle

\begin{abstract}
 Data mining algorithms are now able to efficiently deal with huge amount of data. Various kinds of patterns may be discovered and may have some great impact on the general development of knowledge. In many domains, end users may want to have their data mined by data mining tools in order to extract patterns that could impact their business. Nevertheless, those users are often overwhelmed by the large quantity of patterns extracted in such a situation. Moreover, some privacy issues, or some commercial one may lead the users not to be able to mine the data by themselves. Thus, the users may not have the possibility to perform many experiments integrating various constraints in order to focus on specific patterns they would like to extract. Post processing of patterns may be an answer to that drawback. Thus, in this paper we present a framework that could allow end users to manage collections of patterns. We propose to use an efficient data structure on which some algebraic operators may be used in order to retrieve or access patterns in pattern bases.
\end{abstract}


\section{Introduction}

The amount of information that has been stored in data bases all around the world has continously increased among the years. In order to explore these potential mines of knowledge, efficient data mining tools have been designed for many years. Hence, it is now possible to mine huge databases in order to extract various kinds of patterns, modeling some knowledge. Depending on the algorithms used by end users for their needs, patterns may be varied, we may cite for example decision trees, association rules, formal concepts, etc. While mining huge databases is becoming a common task for many users, those one are now faced with a new problem: how can they exploit the large amount of patterns that are commonly extracted by the data mining tools. Indeed, in the same way it was impossible to manually extract knowledge from huge databases, it is now impossible to manage large volumes of patterns and the end users are in need of new tools in order to do that.

In fact two approaches have been proposed to users in order to manage and explore what is commonly called Pattern Bases. The first one is based on the concept of inductive databases \cite{imielinskietal,boulicautetal99a,djl02,DeRaedt2002}. In Europe, the CInQ project\footnote{http://www.cinq-project.org/} has played a dynamic role in researches in that domain. An inductive
database not only contains data but also patterns and data mining
languages integrated in the inductive database management systems offer some facilities
for pattern manipulation through post-processing operators
\cite{Boulicaut2005b}. Nevertheless those one are very basic and pattern
base management systems should provide more sophisticated functionalities.\\

The second approach for managing patterns focuses on Pattern Base Management Systems (PBMS). In \cite{Catania2004,Catania2006}, a PBMS is defined as "a system for handling (storing / processing / retrieving) patterns defined over raw data in order to efficiently support pattern matching and to exploit pattern-related operations generating intentional information". Thus, the principle consists in storing the patterns extracted by some data mining systems using some efficient data structures. Pattern manipulation
languages have then to be designed in order to manage them. This approach involves two questions. The first one concerns the possibility to design a generic model for patterns, the second one concerns the language needed to access and query patterns. The PANDA project\footnote{http://dke.cti.gr/panda} \cite{pandaproject} is an interesting work in that way. It proposes a generic framework to model various classes of patterns, then
some SQL-like operators allow the user to manage them. Nevertheless, as the underlying model used for storing the patterns is the relational model, the requests that can be designed by users are very complex, non intuitive and time consuming. Even if SQL may be considered an obvious candidate to manage collections of patterns, it was in fact designed to access data stored in databases and it is not well suited to manage patterns \cite{Parsaye99}.
Zaki also proposed in \cite{Zaki2005b} a generic framework for specifying data structures and management functionalities on patterns. Tuzhilin \cite{Tuzhilin2002} specifies some SQL-like operators in order to explore sets of association rules. In those two cases, while some efforts have been done in order to efficiently store patterns, the languages proposed to handle them are quite poor. Finally, in the field of pattern base management, we may cite the PMML project \cite{Grossman99pmml} that allows interoperability of pattern bases, specifying an XML framework
associated to the concept of pattern. Nevertheless this framework is more concerned with structured  representation of patterns than with their management.\\

Our work also belongs to this second approach based on the post processing of patterns.
That is we aim at designing a data structure and efficient algorithms for the management of large pattern bases.
We think that it may be interesting for the users to be able to get
various sets of patterns, that could be successively extracted running data mining
tools on various databases, and then to use efficient tools to manage them.
Indeed, in many cases, due to privacy issues or commercial one, the user does not have any access to the data. 
In this paper, we propose a framework for the management of a particular class of patterns that are called concepts~\cite{WIL92}. More precisely, our approach is based on labeled graphs to
represent collections of concepts. In this domain few works have been done.
The most related one to ours is probably the work of Mielikäinen
\cite{mielikainen04automata} who suggested to represent patterns
using deterministic finite automata. The results obtained experimentally
show that minimum automata provide a compact representation. Nevertheless,
Mielikäinen considered collections of itemsets and not of concepts. Moreover he does not provide any generic framework, based on some algebraic operators.

The next section recalls some basic definitions useful for the understanding of the paper. In
Section~3, we introduce the labeled graph representation of concepts
collections while Section~4 presents a basic algorithm to build this graph.
In Section~5 we define operators that allow to query the graph and that
can be combined using an algebra (in some sense, this section is related to~\cite{DiopGLS04}).

\section{Definitions}
\defnt{A database} \db\ is a  relation between a \defnt{set of
  attributes} $\AA = \set{a_1, a_2,...}$ and a \defnt{set of objects}
$\OO=\set{o_1, o_2,...}$.

Such a database can be represented as a boolean matrix where the columns are
attributes and the rows are objects. 

\begin{figure}[ht]
    \centering
    \begin{tabular}{c||*{6}{c|}}
      & A & B & C & D & E & F \\\hline\hline
      1 & 1 & 1 & 0 & 1 & 1 & 1 \\\hline
      2 & 1 & 1 & 1 & 1 & 0 & 0 \\\hline
      3 & 1 & 1 & 0 & 1 & 1 & 0 \\\hline
      4 & 1 & 1 & 0 & 1 & 0 & 1 \\\hline
      5 & 0 & 0 & 1 & 1 & 1 & 0 \\\hline
    \end{tabular}
    \includegraphics[scale = 0.6]{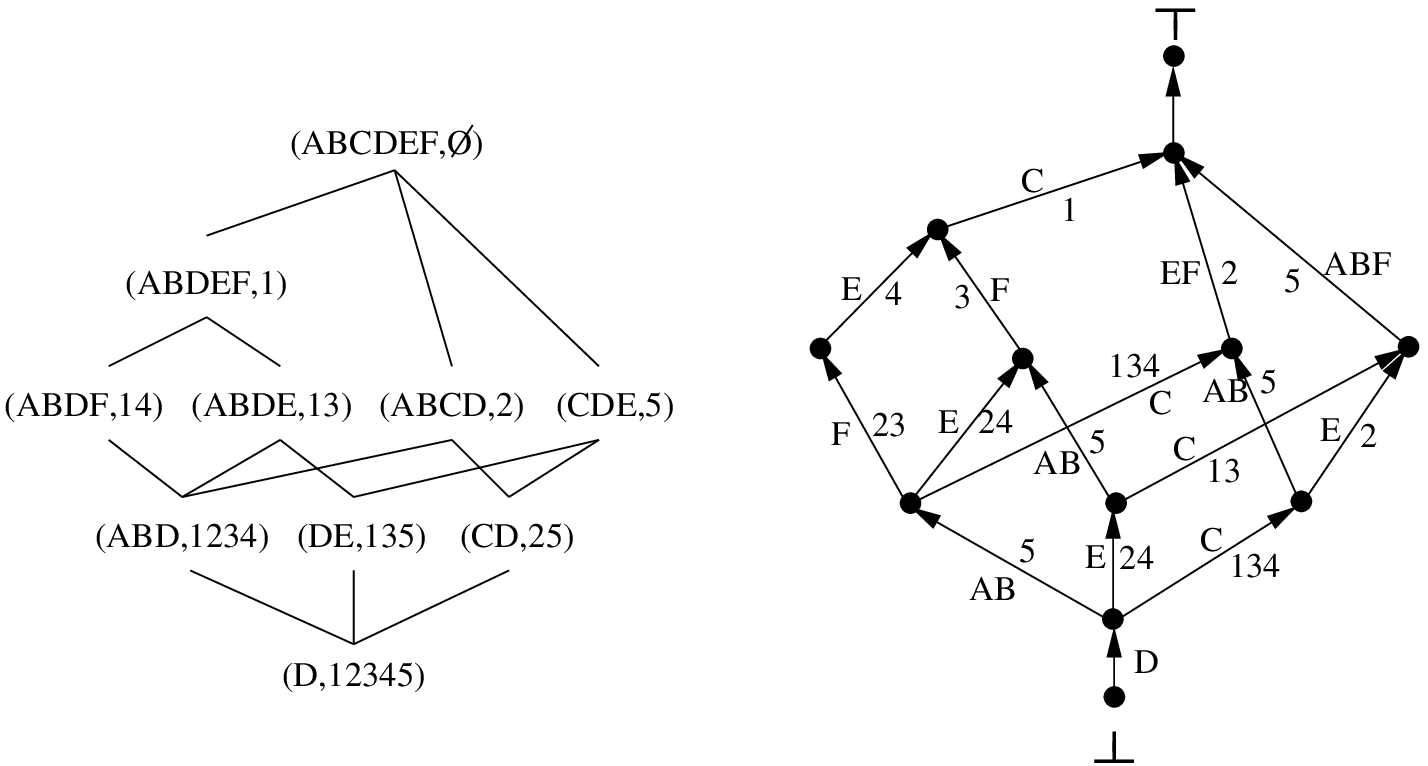}    

  \caption{Example of a database where $\AA=\set{A,B,C,D,E,F}$ and
    $\OO=\set{1,2,3,4,5}$ (top), Hasse diagram of the formal concept
    collection $\concept(\db)$ (left) and the corresponding graph
    representation with labels on the edges (right, see
    Sect.~\ref{sec:repr-conc-coll})}
  \label{fig:db_et_concepts}
\end{figure}

For instance, this database can be the result of gene expression measures. In 
this case, the columns represent genes and the rows represent biological 
 situations. There is a relation between a gene and a situation if the gene
is over-expressed in the given situation. Mining formal concepts in this
kind of data has been shown to be interesting for 
biologists~\cite{idaBessonRBR05}.

A \defnt{bi-set} is a pair $(X,Y)$ where $X\sseq \AA$ and $Y \sseq \OO$.
A \defnt{1-rectangle} is a bi-set $(X,Y)$ such that all the attributes of
$X$ are in relation with all the objects of $Y$. In the matrix, a 1-rectangle
thus defines a sub-matrix containing only ones. 

\begin{example}
  In our example of 
  Fig.~\ref{fig:db_et_concepts}, $(ABD,123)$ (we use this notation for 
  $(\set{A,B,D}, \set{1,2,3})$) and $(E,135)$ are 1-rectangles. 
  $(ABC,12)$ is a bi-set but is not a 1-rectangle since $C$ is not in relation
  with $1$. 
\end{example}

The inclusion  $\sseq$ on bi-sets is defined by: 
$(X_1,Y_1) \sseq (X_2,Y_2)$ iff $X_1 \sseq X_2$ and $Y_1 \sseq Y_2$.
A \defnt{formal concept} is then a maximal 1-rectangle for the order defined
on bi-sets by the inclusion. The collection of all formal concepts in
a database \db\ is $\concept(\db)$ (see Fig.~\ref{fig:db_et_concepts}).

We then define an \defnt{order on the concepts} as follow:
$(X,Y) \preceq (X',Y')$ iff $X\sseq X'$
and $Y'\sseq Y$ (notice the direction of the inclusion).
With this order, the collection of formal concepts forms the well known
formal concept lattice. The Hasse diagram of this lattice for our running
example is presented in
Fig.~\ref{fig:db_et_concepts} (left).

\section{Representation of a Collection of Concepts}
\label{sec:repr-conc-coll}

There are several desirable properties for a good representation:
\begin{itemize}
\item The representation must allow querying: for instance, given a
  collection $\CC$ of concepts, we want to be able to select all concepts
  containing a given attribute or object, or all the concepts containing at
  least 5 objects\ldots
\item The result of a query must be a collection of concepts with the same
  representation as the original collection (closure property).  This is
  important to support successive queries on a collection.
\item In the definitions, there is a duality between objects and
  attributes.  The representation should respect this duality. If it is
  the case, we can use "dual" algorithms for dual operations. For
  instance, the algorithm to select all concepts containing a given
  attribute will be the dual of the algorithm selecting all concepts
  containing a given object.
\end{itemize}

 The output of concept extraction algorithms (such as
D-miner~\cite{idaBessonRBR05}) is typically a file containing a list of
concepts. This is probably the most simple way to represent a collection of
concepts.

Mielikäinen~\cite{mielikainen04automata} proposed to use an automaton to
store an itemset collection (an itemset is a set of attributes). Several
automata are possible: for instance a simple prefix tree or a minimum
automaton.  However, it is necessary to define an order on the attributes
to transform itemsets into strings and choosing a good ordering is very
difficult~\cite{mielikainen04automata} and not very natural. 
Using an automaton to represent concepts is also possible if we can transform
concepts into strings. However, doing this without introducing a arbitrary
order or losing the duality between objects and attributes seems very 
difficult. 

To solve the problem of the need to choose an order, Mielikainen proposed
to use what he called commutative automata~\cite{mielikainen04automata}.
However, these automata have a lot of edges and this is an issue if we want
to query efficiently the representation. Furthermore, the commutative
automata only store the attributes of concepts (and not the objects). This
means that the duality is of course lost and that it will be impossible to
query the set of objects of the concepts without recomputing them.


We propose to use a labeled graph: the Hasse diagram of the order $\preceq$
on the collection of concepts: the vertices are the concepts and there is
an edge $X \rightarrow Y$ between the concepts $X$ and $Y$ iff $Y$ cover
$X$, i.e., $X \prec Y$ and it does not exist a concept $Z$ such that $X
\prec Z \prec Y$.
We add two special vertices: $\bot$ and $\top$ such that 
$(X,Y) \prec \top$ and $\bot \prec (X,Y)$ for all $(X,Y)$.

We can choose to put the labels on the edges or on vertices:
On the vertices: the label consists of the two sets $X$ and $Y$.
On the edges: on the edge $(X,Y) \rightarrow (X',Y')$ the label consists of 
the sets $X'\setminus X$  and $Y \setminus Y'$. 

Figure~\ref{fig:db_et_concepts} shows an example of the constructed graph
with the collection of all the concepts in the database.

With this representation, we do not need to order the attributes or the 
objects and we will show that it is easy to query this representation.

\section{Construction of the Graph Representation}
Given a list of concepts extracted by a concept extraction algorithm such
 as D-miner~\cite{idaBessonRBR05}, the following algorithm constructs the
graph representing the collection. In fact this algorithm is a common release 
of classical algorithms that have been investigated by the Formal Concept Analysis community \cite{Wille1999} in order to build a graph representation of concepts. As this is not the core of our paper, we do not provide too much details on this construction.

The idea of the construction of the graph is to start from a graph
representing the empty collection (which contains only the vertices $\top$
and $\bot$) and to insert the other concepts in the graph one after the
other. In order to simplify the algorithm, we choose to add the concepts
$(X,Y)$ in order of the increasing size of $X$.

When a new concept $C=(X,Y)$ is inserted, there is no other concept
$C'$ in the graph such that $C \preceq C'$ (because of the order in
which the concepts are inserted). Therefore, the only successor of $C$ is
$\top$ and an arc $C \rightarrow \top$ is
added. Next, we must find all predecessors $C'$ of $C$ in the graph 
(i.e., the concepts $C'$ in the graph such that $C$
covers $C'$) to create the arcs $C' \rightarrow C$. 

For this purpose, a depth
first traversal of the graph is performed (starting from $\top$).
The whole graph does not need to be traversed: each time that a concept
$C'$ covered by $C$ is found, there is no need to explore the concepts
smaller than $C'$ (for $\preceq$) since none of them can be covered by $C$. 

Finally, if $C$ covers a concept $C'$ that was
covered by $\top$, the edge $C' \rightarrow \top$ must  be removed
(since $\top$ no longer covers $C'$).

\newcommand{\insertv}{\texttt{insert\_vertex}}
\newcommand{\inserte}{\texttt{insert\_edge}}
\newcommand{\insertc}{\texttt{insert\_concept}}
\newcommand{\rinsert}{\texttt{rec\_insert}}
\newcommand{\dele}{\texttt{delete\_edge}}

This is implemented by the algorithm \texttt{construct\_graph}.  It uses
functions to manipulate the graph (\insertv, \inserte\ and \dele\
which are not detailed) and call a procedure \texttt{insert\_concept} to
insert the next concept in the graph. This procedure call a recursive
procedure \texttt{rec\_insert} to traverse the graph (the set $E$ is used
to "mark" the vertices that have been explored).

\noindent
  \begin{algorithm}[H]
    \dontprintsemicolon
    \caption{construct\_graph}
    \KwIn{An ordered collection $\CC$ of concepts}
    \KwOut{A graph $G$ representing the collection $\CC$}
    $G =$ empty\_graph\;
    \insertv($\top$, $G$)\;
    \insertv($\bot$, $G$)\;
    \inserte($\bot \rightarrow \top$, $G$)\;
    \ForAll{$B \in \CC$}{
      \insertc($B$, $G$)\;
    }
    \Return $G$\;
  \end{algorithm}
  \footnotesize
  \begin{procedure}[H]
    \dontprintsemicolon
    \caption{insert\_concept(concept $B$, graph $G$)}
    \insertv($B$, $G$)\;
    $E = \emptyset$ \tcp*[f]{$E$ is a global variable} \;
    \ForAll{$X \in \mbox{predecessor}(\top)$}{
      \eIf{$X \preceq B$}{
        \dele($X \rightarrow \top$, $G$)\;
        \inserte($X \rightarrow B$, $G$)\;
      }
      {
        \rinsert($B$, $X$, $G$)\;
      }  
    }
    \inserte($B \rightarrow \top$, $G$)\;  
  \end{procedure}

\begin{procedure}
  \dontprintsemicolon
  \caption{rec\_insert(concept $B$, vertex $V$, graph $G$)}
  \ForAll{$X \in \mbox{predecessor}(V) \setminus E$}{
    $E = E \cup \set X$\;
    \eIf{$X \preceq B$}{
      \If{$\nexists Y \in \mbox{successor(X)} \mbox{ such that }  Y\preceq B$}{
        \inserte($X \rightarrow B$, $G$)\;
      }
    }
    {
      \rinsert($B$, $X$, $G$)\;
    }  
  }
\end{procedure}

\section{Queries}
In this section, we study different operations that can be made on a
collection of concepts. We distinguish two kind of queries: selection and
projection queries.

\begin{figure*}[htb]
  \centering
  \includegraphics[scale=0.7]{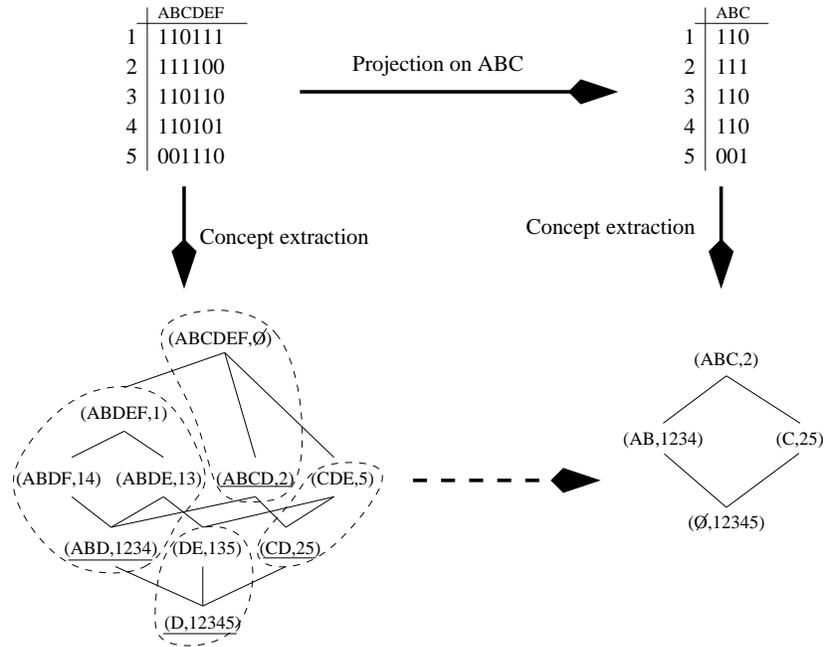}
  \caption{Projection of concepts. The original database $\db$ and the
    corresponding concepts $\concept(\db)$ (left); The projected database
    $\pi_{\set{A,B,C}}(\db)$ and $concept(\pi_{\set{A,B,C}}(\db))$ (right).
    The $\set{A,B,C}$-equivalence classes (dotted, see
    Def.~\ref{def:A_equ}) and their least elements (underlined).  One can
    check the fact that the intersection of the least elements with
    $\set{A,B,C}$ are exactly the concepts of $\pi_{\set{A,B,C}}(\db)$
    (Theorem~\ref{th:projection}).}
  \label{fig:proj_concept}
\end{figure*}

\subsection{Selection Queries}

Given a collection of concepts $\CC$ and a predicate $p$ on the concepts,
we define the selection with respect to $p$ as
$$ \sigma_p(C) = \set{ (X,Y) \in C \mid p(X,Y) \mbox{ is true} }$$

\begin{example}
  Classical examples of selection predicates include~\cite{SouletC05}:
  \begin{itemize}
  \item minimum (or maximum) length: $p(X,Y) = (\size X > \gamma)$
  \item minimum (or maximum) frequency:\\ $p(X,Y) = (\size Y > \gamma)$
  \item minimum (or maximum) area:\\ $p(X,Y) = ( \size X .\size Y > \gamma)$.
  \item requiring that an attribute (object) belongs (does not belong)
    to a concept: $p(X,Y) = (A \in X)$.
  \item \dots
  \end{itemize}
\end{example}

\subsection{Projection Queries}
For example, given gene expression data, a biologist might be interested in
only a part of the genes. He may want to focus only on a
subset of the genes, for instance the genes $A$,$B$ and $C$.

The most simple solution would be to extract the concepts not on the whole
dataset, but on a part of it containing only the columns $A$, $B$ and $C$,
i.e., on a projection of the original database (see
Fig.~\ref{fig:proj_concept}, right). If $A$ is a set of attributes, we
denote $\pi_A(\db)$ the projection of the database $\db$ on the attributes
of $A$.

However, a new extraction of concepts in the projected database would be 
expensive. Furthermore, the original data are perhaps not available anymore (for privacy purposes for example). 
If the collection of concepts in
the whole database is still available, a natural question is whether it
is possible to compute the collection of concepts in the projected database
from the concepts in the whole database (i.e., to find the operation
corresponding to the dotted arrow in Fig.~\ref{fig:proj_concept}).

\begin{sloppypar}
In other words, we want to be able to compute $\concept(\pi_A(\db))$ from
$\concept(\db)$ without having to perform an extraction in $\pi_A(\db)$.
It is indeed possible. First, we need to define an $A$-equivalence relation
on the concepts. 
\end{sloppypar}

\begin{definition}[$A$-equivalence]
  \label{def:A_equ}
  Given a set $A$ of attributes, two concepts $(X,Y)$ and $(X',Y')$ are 
  $A$-equivalent iff $X\cap A= X'\cap A$.
\end{definition}

This is obviously an equivalence relation. Figure~\ref{fig:proj_concept}
gives an example of the equivalence classes. Furthermore, we have the
following proposition:

\begin{proposition}
  The $A$-equivalence classes have a least element (for $\preceq$).
\end{proposition}

To prove this proposition, we use the following well known result: if
$C_1=(X_1,Y_1)$ and $C_2=(X_2,Y_2)$ are two concepts, then there exists a
concept $C=(X_1 \cap X_2,Y)$ with $Y_1 \cup Y_2 \sseq Y$.

\begin{proof}
  Given two $A$-equivalent concepts $C_1=(X_1,Y_1)$ and $C_2=(X_2,Y_2)$, 
  then there exists a third concept $C=(X_1 \cap X_2,Y)$ with 
  $Y_1 \cup Y_2 \sseq Y$.

  Of course, $C$ is $A$-equivalent to $C_1$ and $C_2$ and we also have
  $C \preceq C_1$ and $C \preceq C_2$ (by definition of $\preceq$).

  Therefore, the $A$-equivalence class of $C_1$ and $C_2$ has only
  one minimum element, i.e., it has a least element.

  \qed
\end{proof}

\begin{sloppypar}
The following theorem characterizes the collection $\concept(\pi_A(\db))$
with respect to $\concept(\db)$.   
\end{sloppypar}

\begin{theorem}
  \label{th:projection}
  Given a database $\db$ and a set of attributes $A$, we denote by
  $\mathrm{LE}_A$ the set of the least elements of the $A$-equivalence
  classes.  Then
  $$\concept(\pi_A(\db)) = \set{(X\cap A,Y) \mid (X,Y) \in \mathrm{LE}_A}.$$
\end{theorem}

\begin{proof}
  In this proof, we use the fact that if $(X,Y)$ is a concept in
  $\pi_A(\db)$ then it can be "extended" to form a concept $(X',Y)$ in
  $\db$ where
  $X' \cap A = X$.\\
  First inclusion $\sseq$ :\\
  Let $(X,Y)$ be a concept in $\pi_A(\db)$.  We can "extend" it to a
  concept $(X',Y)$ of $\db$. Let $(X'',Y'')$ be a concept $A$-equivalent to
  $(X',Y)$ such that $(X'',Y'') \preceq (X',Y)$.  $(X''\cap A,
  Y'')=(X,Y'')$ is a 1-rectangle of $\pi_A(\db)$. Since $Y \sseq Y''$ and
  $(X,Y)$ is a concept of $\pi_A(\db)$, $Y$ and $Y''$ are equal. Therefore
  $(X'',Y'')$ is included in $(X',Y)$ and therefore $X'' = X'$ which means
  that $(X'',Y'') = (X',Y)$ and
  $(X',Y)$ is the least element of its $A$-equivalence class.\\
  Inclusion $\supseteq$ :\\
  Let $(X,Y) \in\mathrm{LE}_A $. Then $(X\cap A,Y)$ is a 1-rectangle in
  $\pi_A(\db)$. Suppose that there exists a 1-rectangle $(X',Y')$ in
  $\pi_A(\db)$ such that $(X',Y') \supseteq (X\cap A, Y)$.  Then $X' = X
  \cap A$ otherwise $(X \cup X',Y)$ is a 1-rectangle strictly containing
  $(X,Y)$ and therefore $(X,Y)$ cannot be a concept.  We can extend
  $(X',Y')=(X \cap A, Y')$ to a concept $(X'',Y')$ of $\db$.  Then
  $X''\sseq X$ otherwise $(X\cup X'',Y)$ is a 1-rectangle strictly
  containing $(X,Y)$ and thus $(X,Y)$ cannot be a concept. Therefore
  $(X'',Y') \preceq (X,Y)$ and these two concepts are $A$-equivalent.
  Therefore they are equal (since $(X,Y)$ is a least element) and $(X\cap
  A,Y)$ is a maximal 1-rectangle in $\pi_A(\db)$ (for $\sseq$), i.e., a
  concept of $\pi_A(\db)$.  \qed
\end{proof}

More generally, we can define a projection operation on collections of 
concepts:

\begin{definition}[collection of concepts projection]
  \label{def:proj}
  \begin{sloppypar}
  Given a collection of concepts $\CC$ in a database $\db$ and a set of
  attributes $A$, we define the projection of the collection $\CC$ with respect
  to $A$ by:
  $$ \pi_A(\CC) = \set{(X\cap A, Y) \mid (X,Y) \in \CC \cap \mathrm{LE}_A}$$
  where $\mathrm{LE}_A$ is defined as in Theorem~\ref{th:projection}.
  \end{sloppypar}
\end{definition}

Theorem~\ref{th:projection} 
means that this projection operation can be used to compute
the concepts in the projected database $\pi_A(\db)$ by projecting
the concepts of the original database $\db$: 
$\concept(\pi_A(\db)) = \pi_A(\concept(\db))$. In this equality,
the first $\pi_A$ denotes a database projection whereas the second one
denotes a collection of concepts projection (Def.~\ref{def:proj}).

\subsection{Algebra}

In this section, we study how the projection and selection operations on
collection of concepts compose with each other.

We want to know if
there exists an operation to close the following diagram (dotted arrow).
A natural candidate is the projection that we have just defined.\\
\begin{center}
  \includegraphics[scale=0.7]{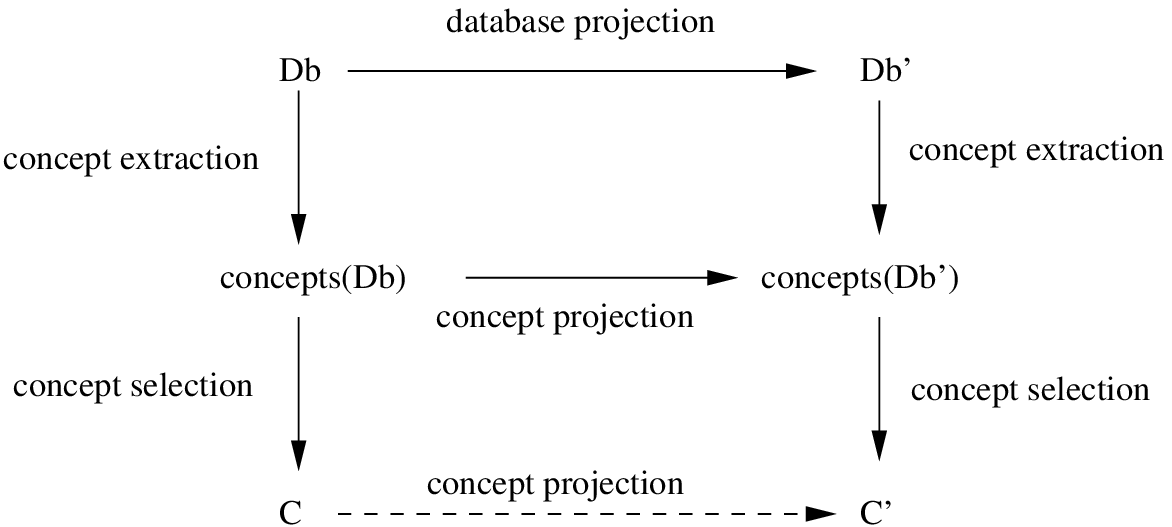}
\end{center}

Indeed, the following theorem shows that this diagram can be closed using the
projection operation:

\begin{theorem}
  \label{th:sel_proj}
  Given a collection of concepts $\CC$ in a database $\db$, a set of attributes $A$
  and a selection predicate $p$ such that for all concepts $(X,Y)$, $p(X\cap
  A, Y) = p(X,Y)$, then
  $$ \pi_A\circ \sigma_p (\CC)  = \sigma_p \circ \pi_A (\CC).$$
\end{theorem}

\newcommand{\equivalent}{\Longleftrightarrow}
\begin{proof}
  $(X,Y) \in \pi_A(\sigma_p(\CC))$
  $\equivalent$
  $\exists (X',Y) \in \sigma_p(\CC) \cap \mathrm{LE}_A$ such that
  $X = X' \cap A$ (by Def.~\ref{def:proj})
  $\equivalent$
  $\exists (X',Y) \in \mathrm{LE}_A \cap \CC$ such that
  $p(X',Y)$ is true and $X = X' \cap A$ 
  $\equivalent$
  $\exists (X',Y) \in \mathrm{LE}_A \cap \CC$ such that
  $p(X,Y)$ is true and $X = X' \cap A$ 
  (since $p(X', Y) = p(X'\cap A,Y) = p(X,Y)$)
  $\equivalent$
  $(X,Y) \in \pi_A(\CC)$ and $p(X,Y)$ is true (by Def.~\ref{def:proj})
  $\equivalent$
  $(X,Y) \in \sigma_p(\pi_A(\CC))$
  \qed
\end{proof}

The requirement on $p$ can seem very strong but it is necessary. In order
to be able to perform the selection after the projection, the projection
must not remove too much information from the collection. For instance, if
the selection is defined by $p(X,Y) = (D \in X)$ (i.e., select the concepts
containing attribute $D$), then this selection does not commute with the
projection $\pi_{\set{A,B,C}}$. Indeed, after this projection the
information whether a concept contained attribute $D$ is no longer
available. There is a similar behavior with selection and projection
defined on relational tables. If the selection uses an attribute
which is suppressed by the projection, the two operations do not commute.

\subsection{Duality}
In the two previous sections, we defined the projection of a collection of
concepts on a set $A$ of attributes. In a dual manner, we can define another
projection on a set $O$ of objects. The dual equivalence relation of the
$A$-equivalence (Def.~\ref{def:A_equ}) can be defined as follow: two
concepts $(X,Y)$ and $(X',Y')$ are $O$-equivalent iff $Y\cap O = Y'\cap O$.
Then we have the dual of theorems \ref{th:projection}~and~\ref{th:sel_proj}.

\subsection{Algorithms}

In this section, we present the algorithms to actually perform the
selection and projection on the graph representation of the collection.

To perform the projection of a collection of concepts $\CC$ with respect to
a set of attributes $A$, we need to be able to test if a concept is minimal
in its $A$-equivalence class. However, this is not always possible without
additional information: it is possible that the collection $\CC$ does not
contain all the concepts belonging to an equivalence class, in this case,
we could find a minimum concept in this equivalence class in $\CC$ which is
not the least element of this equivalence class in $\concept(\db)$.

For instance, suppose $\CC$ contains all the concepts of
Fig.~\ref{fig:proj_concept} (before projection) except concept $(D,12345)$.
Then, if we compute the projection of this collection with respect to
$\set{A,B,C}$, we must be able to detect that $(DE,135)$ is not a
least element of an equivalence class. Without additional information, it
is not possible without a possibly expensive check in the data.

\newcommand{\insertmc}{\texttt{insert\_marked\_vertex}}
This is the reason why we add some information in our graph representation.
Given a collection $\CC$ of concepts, we add into the graph the concepts
that are "just outside" of the collection. By just outside, we mean the
concepts that are either predecessor or successor of a concept belonging to
the collection. These additional concepts are marked and are not linked to
$\top$ and $\bot$ (they are inserted in the graph with the \insertmc\
function), they are linked only to the concept(s) of the collection which
is (are) their predecessor or successor. Of course, when doing selection or
projection operations, this additional information must be maintained.

The algorithm to perform the projection is given in Alg.~\ref{alg:proj}.
For all vertex $X$, the algorithm computes \texttt{le[X]} which is the least
element of the $A$-equivalence class of $X$. If this least element is not
in the collection, then \texttt{le[X]=NIL}.  The least elements of the
equivalence classes are inserted in the new graph $G'$ and the edges are
added to $G'$.  In the algorithm, we use the notation \texttt{proj(X,Y)} to
denote $(X\cap A, Y)$.

  \begin{algorithm}[H]
    \label{alg:sel_am}
    \dontprintsemicolon
    \caption{selection\_AM}
    \KwIn{A graph $G$ representing a collection $\CC$ of concepts and an
      anti-monotonic selection predicate $p$}
    \KwOut{A graph $G'$ representing the collection $\sigma_p(\CC)$}
    
    $G' =$ empty\_graph\;
    \insertv($\top$, $G'$)\;
    \insertv($\bot$, $G'$)\;
    $E = \emptyset$ \tcp*[f]{$E$ is a global variable}\;
    explore($\bot$)\;
    return $G'$\;
  \end{algorithm}

In the general case, to compute the selection of a collection of concepts
with respect to a predicate $p$, we must traverse the graph representing
the collection and test $p$ on all concepts. 

\begin{sloppypar}
  
However, when $p$ is monotonic or anti-monotonic, it is not necessary to
traverse the whole graph.  A predicate $p$ is anti-monotonic iff $ (\neg
p(X,Y) \et ((X,Y) \preceq (X',Y'))) \Rightarrow \neg p(X',Y') $ and
monotonic iff $ (\neg p(X,Y) \et ((X',Y') \preceq (X,Y))) \Rightarrow \neg
p(X',Y') $.  Therefore, if $p$ is anti-monotonic, the graph can be explored
bottom up (from $\bot$ to $\top$) and if a concept $X$ that does not
satisfy $p$ is found, it is not necessary to explore its successors (see
Alg.~\ref{alg:sel_am}).  Dually, for a monotonic constraint, the graph is
explored top down.

\end{sloppypar}

  \begin{procedure}[H]
    \dontprintsemicolon
    \caption{explore(vertex V)}
    $E = E \cup \set V$ \tcp*[f]{$E$ is a global variable}\;
    \ForAll{$X \in \mbox{predecessor}(V)$ and $X$ marked}{
      \insertmc($X$, $G'$)\;
      \inserte($X \rightarrow V$, $G'$)\;
    }
    link\_to\_top = true\;
    \ForAll{$X \in \mbox{successor}(V)$}{
      \eIf{$p(X)$ and $X$ not marked}{
        link\_to\_top = false\;
        \If{$X \not \in E$}{
          \insertv($X$, $G'$)\;
          explore($X$)\;
        }
        \inserte($V \rightarrow X$, $G'$)\;
      }
      {
        \insertmc($X$,$G'$)\;
        \inserte($V \rightarrow X$, $G'$)\;
      }
    }
    \If{link\_to\_top}{
      \inserte($V \rightarrow \top$)\;
    }
  \end{procedure}
\begin{algorithm}
  \footnotesize
  \label{alg:proj}
  \dontprintsemicolon
  \caption{projection}
  \KwIn{A graph $G$ representing a collection $\CC$ of concepts and a
    set  $A$ of attributes}
  \KwOut{A graph $G'$ representing the collection $\pi_A(\CC)$}
  \ForAll{$X \in \CC$, $X$ not marked}{
    le[$X$]= $X$\;
  }
  \ForAll{$X \in \CC$, $X$ not marked, in topological order}{
    \If(\tcp*[f]{$X$ is perhaps in $\mathrm{LE}_A$}){le[$X$] = $X$}{
      \eIf(\tcp*[f]{$X$ is not in $\mathrm{LE}_A$}){$\exists Y \in predecessor(X)$, $Y$ marked and class($Y$) = class($X$)}{
        le[$X$] = NIL \;
      }(\tcp*[f]{$X$ is in $\mathrm{LE}_A$})
      {\insertv(proj($X$), $G'$)\;
        \ForAll{$X'$ marked $\in predecessor(X)$}{
          \insertmc(proj($X'$), $G'$)\;
          \inserte($proj(X') \rightarrow proj(X)$, $G'$)\;
        }
      }
    }
    \ForAll{ $Y$ unmarked $\in successor(X)$}{
      \If{class($Y$) = class($X$)}{
        le[$Y$] = le[$X$]\;
      }
    }
  }
  \ForAll{edge $X \rightarrow Y$ in $G$, $X$ and $Y$ unmarked}{
    \inserte(proj(le[$X$]) $\rightarrow$ proj(le[$Y$]), $G'$)\;
  }
\end{algorithm}

\section{Conclusion}

In this article, we made an original study on how to represent and query
collections of concepts.  We proposed to store these
collections using a graph representation and we defined two kinds of operators:
selection and projection. 

We want to extend this work in several directions. First, it would be interesting to study the scalability of our  representation on real datasets and make comparison with, for instance, automata representations. Studying the relationships between the size of the representation and the characteristics of the datasets from which it was extracted would also be interesting.

Second, our representation using a graph is efficient for querying but is could be more compact. One could use two representations of the collection of concepts: a very compact one for long term storage (on disk) and another one (the graph) for querying.

Finally, several works have been done on generalization of concepts and on clustering of concepts. It would be interesting to study if it is  possible to define an aggregation operator (a kind of ``group by'' operator) on the graph to support these generalization facilities..

\section{Acknowledgments}
This work is partially funded by the french ACI "masse de données" (Bingo
project).

\end{document}